\documentclass[final]{cvpr}

\usepackage{times}
\usepackage{epsfig}
\usepackage{graphicx}
\usepackage{amsmath}
\usepackage{amssymb}
\usepackage{mathtools}

\usepackage{format/math_macro} 
\usepackage{enumitem}

\usepackage{soul}

\usepackage{amssymb} 
\usepackage{amsmath} 
\usepackage{booktabs}
\usepackage{multirow}

\newcommand{\lsparagraph}[1]{\vspace{0.1cm} \noindent \textbf{#1}}

\DeclareMathOperator*{\argmin}{arg\,min}

\usepackage{xcolor}
\definecolor{cb-black}      {RGB}{  0,   0,   0}
\definecolor{cb-blue-green} {RGB}{  0,  073,  073}
\definecolor{cb-green-sea}  {RGB}{  0, 146, 146}
\definecolor{cb-rose}       {RGB}{255, 109, 182}
\definecolor{cb-salmon-pink}{RGB}{255, 182, 119}
\definecolor{cb-purple}     {RGB}{ 73,   0, 146}
\definecolor{cb-blue}       {RGB}{ 0, 109, 219}
\definecolor{cb-lilac}      {RGB}{182, 109, 255}
\definecolor{cb-blue-sky}   {RGB}{109, 182, 255}
\definecolor{cb-blue-light} {RGB}{182, 219, 255}
\definecolor{cb-burgundy}   {RGB}{146,   0,   0}
\definecolor{cb-brown}      {RGB}{146,  73,   0}
\definecolor{cb-clay}       {RGB}{219, 209,   0}
\definecolor{cb-green-lime} {RGB}{ 36, 255,  36}
\definecolor{cb-yellow}     {RGB}{255, 255, 109}

\usepackage{pifont}
\newcommand{\cmark}{{\ding{51}}}
\newcommand{\xmark}{}

\usepackage[pagebackref=true,breaklinks=true,colorlinks,bookmarks=false]{hyperref}

\def\cvprPaperID{414} 
\def\confYear{CVPR 2021}

\begin{document}

\title{
UnsupervisedR\&R: \\
Unsupervised Point Cloud Registration via Differentiable Rendering
}

\author{{Mohamed El Banani \qquad Luya Gao \qquad Justin Johnson} \\
University of Michigan\\
{\tt\small \{mbanani, mlgao, justincj\}@umich.edu}
}

\maketitle
\global\csname @topnum\endcsname 0
\global\csname @botnum\endcsname 0

\begin{abstract}

Aligning partial views of a scene into a single whole is essential to understanding one's environment and is a key component of numerous robotics tasks such as SLAM and SfM.
Recent approaches have proposed end-to-end systems that can outperform traditional methods by leveraging pose supervision.
However, with the rising prevalence of cameras with depth sensors, we can expect a new stream of raw RGB-D data without the annotations needed for supervision.
We propose UnsupervisedR\&R: an end-to-end unsupervised approach to learning point cloud registration from raw RGB-D video. 
The key idea is to leverage differentiable alignment and rendering to enforce photometric and geometric consistency between frames.
We evaluate our approach on indoor scene datasets and find that we outperform existing traditional approaches with classic and learned descriptors while being competitive with supervised geometric point cloud registration approaches.

\end{abstract}

\begin{figure}
\begin{center}
   \includegraphics[width=\linewidth]{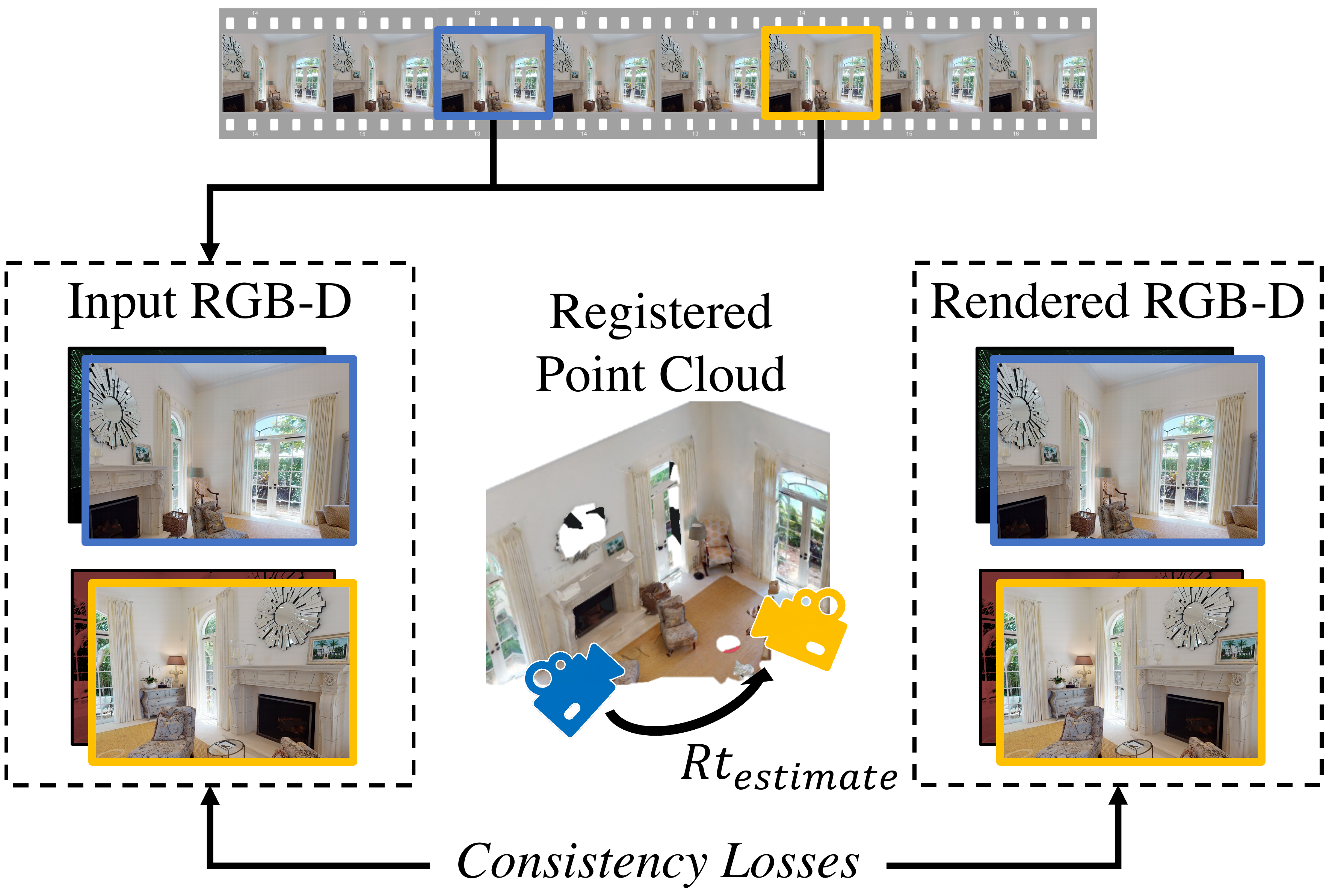}
\end{center}
    \caption{What 3D scene do the two views on the left portray? Given 2 RGB-D images, we train a model to estimate the camera motion between them through enforcing photometric and geometric consistency losses on point cloud renderings of the scene.}
\label{fig:teaser}
\end{figure}

\section{Introduction}
Consider the two scenes depicted in Fig \ref{fig:teaser}. How are they related? What is the layout of the room they depict?
Aligning partial views of a scene into a single whole is essential to understanding one's environment and is a key component of numerous robotics tasks such as SLAM and SfM.
Recent approaches have leveraged supervised learning to develop end-to-end systems that outperform traditional methods in both accuracy and speed~\cite{choy2020deep,gojcic2020learning}.
However, with the rising prevalence of cameras with depth sensors, we can expect a new stream of raw RGB-D data without the annotations needed for supervision.
\textit{How can we leverage this data for unsupervised learning of point cloud registration?}

The common approach to point cloud registration relies on correspondence extraction and geometric model fitting.
Traditional approaches relied on hand-crafted features~\cite{johnson1999using,lowe2004distinctive} and robust estimators such as RANSAC~\cite{RANSAC}.
While those approaches work well, their performance is limited by their inability to flexibly adapt to different data distributions. 
Recent work leverages supervised learning to address those limitations by learning to extract feature descriptors~\cite{choy2019fully,detone2018superpoint,yi2016lift}, 
finding better correspondences~\cite{choy2020deep,gojcic2020learning,sarlin2020superglue}, 
and training more efficient robust estimators~\cite{brachmann2017dsac,brachmann2019neural,ranftl2018deepfundamental}.
However, accurate pose annotation can be challenging to attain automatically, due to sensor error or reliance on traditional SfM pipelines with no convergence guarantees~\cite{schonberger2016structure}. 

Meanwhile, self-supervised visual learning has made remarkable progress in learning semantic ~\cite{desai2020virtex,doersch2015unsupervised,gidaris2018unsupervised,goyal2019scaling,tian2019contrastive} and 3D~\cite{insafutdinov2018pointclouds,kulkarni2020acsm,tulsiani2018multiviewconsistency,ummenhofer2017demon,zhou2017unsupervised} features. 
The key idea is to use natural transformations in the data as indirect supervision. 
RGB-D video provides us with this supervision since successive frames capture different views of the same scene.
In this case, aligning two point clouds from nearby frames is not only about achieving good geometric consistency, but also showing good photometric consistency between the two views. 
By achieving both photometric and geometric consistency, we can train a system using RGB-D image pairs without relying on additional supervision.

We propose using view synthesis between RGB-D images as a task for learning point cloud registration. 
Given two RGB-D video frames, we extract features from each frame to generate a feature point cloud where each point is represented by both a 3D coordinate and a feature vector.
The extracted features serve as descriptors for correspondence estimation.
The model is trained end-to-end using photometric and geometric consistency losses between the input and rendered frames.
Through using differentiable components, we back-propagate the losses to the feature encoder to learn features that allow us to estimate unique correspondences and accurately register the two views.

We evaluate our model on ScanNet~\cite{dai2017scannet}; a large indoor scene dataset. 
We find that our model outperforms the traditional registration pipeline with visual or geometric descriptors (\S~\ref{sec:exp_pcreg}).
Furthermore, it performs on-par with supervised geometric registration approaches despite being unsupervised;
supporting our claim that RGB-D self-supervision can alleviate the need for pose annotation. 
Finally, we analyze our model through several ablations (\S~\ref{sec:exp_ablations}).

In summary, our contributions are as follows:
\begin{itemize}[nosep, leftmargin=*]
    \item We propose an unsupervised approach to point cloud registration via differentiable alignment and rendering;
    \item We show how a differentiable variant of Lowe's ratio test is sufficient for correspondence matching; 
    \item We empirically demonstrate our approach's efficacy against traditional \& supervised registration approaches;
    \item We validate our design choices by evaluating our model with several ablations. 
\end{itemize}

\section{Related Work}
\label{sec:related}
\lsparagraph{Feature Descriptors.}
Early work on feature point extraction can be traced back to using corners for stereo matching~\cite{moravec1981rover}. 
This work culminated in patch-based feature 2D descriptors~\cite{bay2006surf,lowe2004distinctive,rublee2011orb} and geometric features based on histograms of local 3D relationships~\cite{johnson1999using,rusu2009fast}. 
Those descriptors have been very popular due to being efficient to compute, relatively robust, and data-agnostic.
More recently, there has been an interest in leveraging convolutional neural networks to extract good visual descriptors~\cite{detone2018superpoint,dusmanu2019d2,han2015matchnet,yi2016lift,zagoruyko2015learning} and geometric descriptors~\cite{bai2020d3feat,choy20163d,deng2018ppf,deng20193d,gojcic2019perfect,Li2020e2e3ddescriptors,wang2019dcp,yew20183dfeatnet}.
Relevant to our work are approaches that use geometric transformations to learn visual features.
This has been commonly done by using known pose or correspondences between large collections of images~\cite{detone2018superpoint,dusmanu2019d2,yi2016lift} or point clouds~\cite{choy20163d,deng2018ppf,deng20193d,gojcic2019perfect,wang2019dcp}.
We extend this work by using existing transformation in RGB-D video data and relying on consistency losses instead of pose supervision.

\lsparagraph{Correspondence Estimation and Fitting.} 
Early work on image and point cloud registration assume perfect correspondences~\cite{arun1987least,longuet1981computer}.
ICP relaxes this assumption for closely aligned points by introducing the simple heuristic of assuming the closest point is the correspondence~\cite{zhang1994iterative}. 
However, extending to real-world settings requires the ability to determine such correspondences from the raw input or extracted features. 
Early work uses feature similarity and heuristic approaches to determine correspondence and robust estimators such as RANSAC to handle noise and outliers in the correspondences~\cite{lowe2004distinctive, torr1997development, zhang1995robust}. For a review, see ~\cite{pomerleau2015review}. 
More recent approaches advance this idea by learning differentiable functions for weighting the correspondences~\cite{brachmann2017dsac, brachmann2019neural,choy2020deep,gojcic2020learning,huang2020featureregistration,lu2019deepvcp,ranftl2018deepfundamental,sarlin2020superglue,yew2020RPMNet}.
Finally, there have been recent self-supervised approaches for registering object point clouds~\cite{aoki2019pointnetlk,hertz2020pointgmm,huang2020featureregistration,wang2019dcp,wang2019prnet,yew2020RPMNet,yuan2020deepgmr}. Those approaches operate on dense point clouds that are either augmented and sampled for partial views with known pose and correspondences. Hence, while the setup might be self-supervised, the methods still require ground-truth annotation.  
We are inspired by this line of work, but differ from it in two key ways: (1) we take RGB-D images as input, not keypoints and descriptors or 3D scenes; (2) our approach is unsupervised, while those approaches require pose or correspondence supervision.

\lsparagraph{Differentiable SfM.}
There has been a large number of recent approaches that replace the traditional SfM pipeline with end-to-end learning approaches~\cite{casser2019depthprediction, Mahjourian2018depthicp,qian2020associative3d,tang2018banet,teed2019deepv2d,ummenhofer2017demon,vijayanarasimhan17sfmnet,yin2018geonet,zhou2017unsupervised}.
Related to our work are approaches that propose unsupervised learning of depth and camera motion. 
This is typically done through learning two CNNs: a pose network and a depth network, that are trained to minimize a consistency loss between video frames. 
While CNN pose estimators have shown a lot of success on outdoor scenes, they have been challenged by cases with larger and more erratic camera motions (\eg video from a hand-held device)~\cite{bian2020depth}.
Similar to those approaches, we train an end-to-end system using photometric and geometric consistency losses.
Unlike that work, we are interested in pointcloud registration with larger camera motions and learn features for correspondence alignment of RGB-D scans. 

\lsparagraph{View Synthesis.}
View synthesis is the task of generating views of the scene from image inputs.
One line of work focuses on synthesizing views with small camera motions~\cite{kalantari2016learning, niklaus20193d, penner2017soft, shin20193d, srinivasan2019pushing, srinivasan2017learning}. 
NeRF and its variants~\cite{martin2020nerf,mildenhall2020nerf,zhang2020nerf++} learn a rendering function for a specific scene from a large collection of multi-view images.
While the goal of that work is highly photo-realistic renderings, we are primarily interested in utilizing view synthesis as a training task to enforce photometric consistency.
Similar to our goals are approaches that synthesize views for unsupervised 3D learning of object shape~\cite{elbanani2020novelviewpoints,insafutdinov2018pointclouds,kar2017learning,tulsiani2015viewpoints} and depth~\cite{casser2019depthprediction, Mahjourian2018depthicp,ummenhofer2017demon,vijayanarasimhan17sfmnet,yin2018geonet,zhou2017unsupervised}.
Closest to our work is Wiles~\etal~\cite{wiles2019synsin} who train a model to for depth estimation and view synthesis with the goal of generating highly photo-realistic views of the scene.
Our work complements this earlier work since we learn pose while they learn depth.

\begin{figure*}
\begin{center}
   \includegraphics[width=\linewidth]{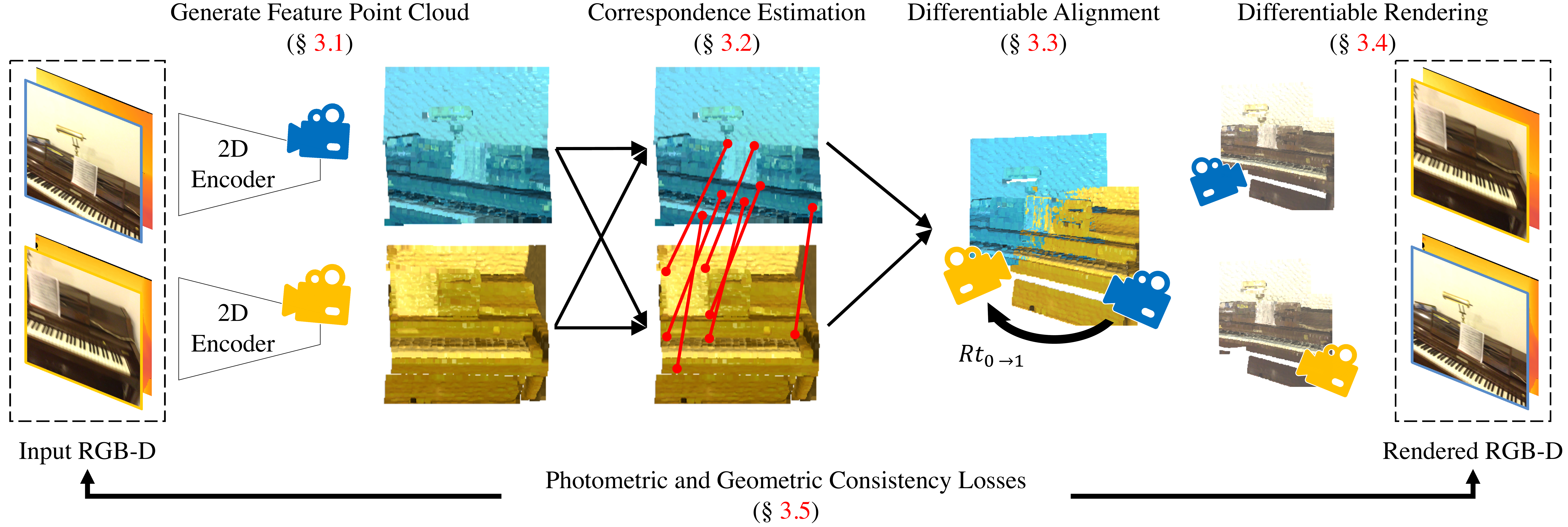}
   \vspace{-0.4cm}
\end{center}
    \caption{\textbf{Our end-to-end approach.} 
    Our model takes as input two RGB-D images of a scene. 
    First, we encode the images into a feature map and project them into a 3D point cloud. 
    Second, we extract correspondences between the two feature point clouds. 
    Third, we use the 3D correspondences to estimate $Rt_{0\to1}$; a 6-DOF transformation that aligns the two point clouds. 
    Finally, we use differentiable rendering to render the points from both point clouds and apply consistency losses.
    }
\label{fig:model}
\end{figure*}

\section{Method}
\label{sec:approach}
The goal of this work is to build a system that can learn point cloud registration from RGB-D video without any supervision.
Our approach, shown in Fig.~\ref{fig:model}, is based on the traditional registration pipeline as it similarly extracts feature descriptors, finds correspondences, and finds the best alignment. 
We adapt this pipeline by operating directly on the images and learning our own features, as well as using photometric and geometric consistency losses to learn those features. 
We first present a high-level sketch of our approach before explaining each stage in more detail. 
Architectural details are presented in the appendix and our code is available at 
{\small \url{https://github.com/mbanani/unsupervisedRR}}.

\paragraph{Approach Sketch. }
Given two RGB-D images of the scene and the camera's intrinsic matrix, we first extract 2D features for each image and project them into two feature point clouds.
We extract correspondences between the two point clouds and rank the correspondences based on their uniqueness.
We then use a differentiable optimizer to align the top $k$ correspondences and estimate the 6-DOF transformation between them. 
Finally, we render the point cloud from the two estimated viewpoints to generate an RGB image for each view. 
We use photometric and geometric consistency losses between the RGB-D inputs and outputs and back-propagate through our entire pipeline. 

\begin{figure*}
\begin{center}
   \includegraphics[width=0.95\linewidth]{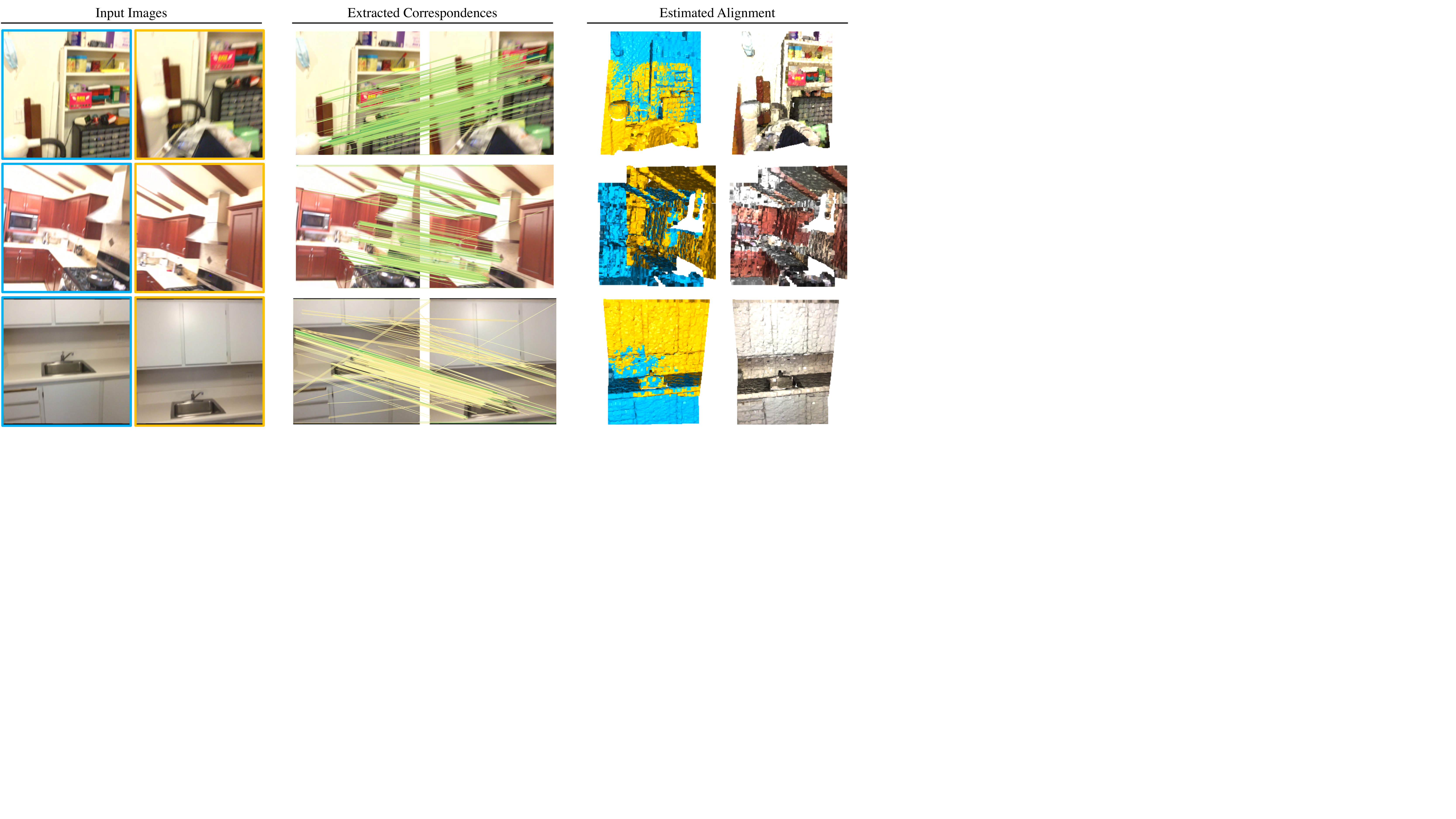}
\end{center}
   \caption{\textbf{Pairwise Registration Results.} Our model extracts dense correspondences and achieves highly accurate alignments for indoor scene datasets. Our correspondences are color-coded using their ratio test weight; green indicates a higher weight.  As shown, for relatively textureless images, the predicted correspondences are less confident, yet our model still able to achieve accurate alignment.}
\label{fig:qualitative}
\end{figure*}

\subsection{Point Cloud Generation}
\label{sec:method_pcgen}

Given an input RGB-D image, $I \in \mathbb{R}^{4 \times H \times W}$, we would like to generate a point cloud $\mathcal{P} \in \mathbb{R}^{(6 + F) \times N}$. 
Each point $p \in \mathcal{P}$ is represented by a 3D coordinate $\mathbf{x}_p \in \mathbb{R}^{3}$, a color $\mathbf{c}_p \in \mathbb{R}^{3}$, and a feature vector $\mathbf{f}_p \in \mathbb{R}^{F}$.
We first use a feature encoder to extract a feature map using each image's RGB channels.
The extracted feature map has the same spatial resolution as the input image. 
As a result, one can easily convert the extracted features and input RGB into a point cloud using the input depth and known camera intrinsic matrix.
However, given that current depth sensors do not predict depth for every pixel, we omit pixels with missing depth from our generated point cloud.
In order to avoid heterogeneous batches, we mark points with missing depths so that subsequent operations ignore them. 

\subsection{Correspondence Estimation}
\label{sec:method_corr}

Given two feature point clouds\footnote{As noted in Sec~\ref{sec:method_pcgen}, point clouds will have different numbers of valid points based on the input depth. While our method deals with this by tracking those points and omitting them from subsequent operations, we assume all the points are valid in our model description to enhance clarity.}, 
$\mathcal{P}$, $\mathcal{Q} \in \mathbb{R}^{(6 + F) \times N}$, 
we would like to find the correspondences between the point clouds.
Specifically, for each point in $p \in \mathcal{P}$, we would like to find the point $q_p$ such that 
\begin{equation}
q_{p} = \operatorname*{arg\,min}_{q\in{\mathcal{Q}}} D(\mathbf{f}_p, \mathbf{f}_q),
\end{equation}
where $D(p, q)$ is a distance-metric defined on the feature space. 
In our experiments, we use cosine distance to determine the closest features. 
 
We extract such correspondences for all the points in both $\mathcal{P}$ and $\mathcal{Q}$ since correspondence is not guaranteed to be bijective. 
As a result, we have two sets of correspondences, $\mathcal{C}_{\mathcal{P} \to \mathcal{Q}}$ and $\mathcal{C}_{\mathcal{Q} \to \mathcal{P}}$, where each set consists of $N$ pairs.

\paragraph{Ratio Test.} 
Determining the quality of each correspondence is a challenge faced by any correspondence-based geometric fitting approach.
Extracting correspondences based on only the nearest neighbor will result in many false positives due to falsely matching repetitive pairs or non-mutually visible portions of the images. 

The standard approach is to estimate a weight for each correspondence that captures the quality of this correspondence.
Recent approaches estimate a correspondence weight for each match using self-attention graph networks~\cite{sarlin2020superglue}, PointNets~\cite{gojcic2020learning,yi2018learning}, and CNNs~\cite{choy2020deep}.
In our experiments, we found that a much simpler approach based on Lowe's ratio test~\cite{lowe2004distinctive} works well without requiring any additional parameters in the network. 
The basic intuition behind the ratio test is that unique correspondences are more likely to be true matches.
As a result, the quality of correspondence $(p, q_p)$ is not simply determined by $D(p, q_p)$, but rather between the ratio $r$ which is defined as 
\begin{equation}
r = \frac{D(p, q_{p, 1})}{D(p, q_{p, 2})},
\end{equation}
where $q_{p, i}$ is the $i$-th nearest neighbor to point $p$ in $\mathcal{Q}$.
Since $0 \leq r_p \leq 1$ and a lower ratio indicates a better match, we weigh each correspondence by $w = 1 - r$.  

In the traditional formulation, one would define a distance ratio threshold for inlier vs outliers. 
Instead, we rank the correspondences using their ratio weight and pick the top $k$ correspondences.
We pick an equal number of correspondences from $\mathcal{C}_{\mathcal{P} \to \mathcal{Q}}$ and $\mathcal{C}_{\mathcal{Q} \to \mathcal{P}}$.
Additionally, we keep the weights for each correspondence to use in the geometric fitting step. 
Hence, we end up with a correspondence set $\mathcal{M} = \{(p, q, w)_i: 0 \leq i < k \}$ where $k{=}400$.

\subsection{Geometric Fitting}
\label{sec:method_fitting}

Given a set of correspondences $\mathcal{M}$, we would like to find the transformation, $\mathcal{T^{*}} \in \text{SE(3)}$ that would minimize the error between the correspondences
\begin{equation}
    \mathcal{T^{*}} = \argmin_{\mathcal{T} \in~\text{SE}(3)} E(\mathcal{M}, \mathcal{T}) 
    \label{eq:w_proc}
\end{equation}
where the error $E(\mathcal{M}, \mathcal{T})$ is defined as:
\begin{equation}
    E(\mathcal{M}, \mathcal{T}) = |\mathcal{M}|^{-1} \sum_{(p, q, w) \in \mathcal{M}} w~(\mathbf{x}_p - \mathcal{T}(\mathbf{x}_q))^2
    \label{eq:corr_err}
\end{equation}
This can be framed as a weighted Procrustes problem and solved using a weighted variant of Kabsch's algorithm~\cite{kabsch1976solution}.

While the original Procrustes problem minimizes the distance between a set of unweighted correspondences~\cite{gower1975generalized}, Choy \etal~\cite{choy2020deep} have shown that one can integrate weights into this optimization.
This is done by calculating the covariance matrix between the centered and weighted point clouds, followed by calculating the SVD on the covariance matrix. 
For more details, see~\cite{choy2020deep, kabsch1976solution}. 

Integrating weights into the optimization is important for two reasons.
First, it allows us to build  robust estimators that can weigh correspondences based on our confidence in their uniqueness. 
More importantly, it makes the optimization differentiable with respect to the weights, allowing us to backpropagate the losses back to the encoder for feature learning. 

\paragraph{Randomized Optimization. }
While this approach is capable of integrating the weights into the optimization, it can still be sensitive to outliers with non-zero weights.
We take inspiration from RANSAC and use random sampling to mitigate the problem of outliers. 
More specifically, we sample $t$ subsets of $\mathcal{M}$, and use Equation~\ref{eq:w_proc} to find $t$ candidate transformations. 
We then choose the candidate that minimizes the weighted error on the full correspondence set. 
Since the $t$ optimizations on the correspondence subsets are all independent, we are able to run them in parallel to make the optimization more efficient. 
We deviate from classic RANSAC pipelines in that we choose the transformation that minimizes a weighted error, instead of maximizing inlier count, to avoid having to define an arbitrary inlier threshold. 

It is worth noting that the model can be trained and tested with a different number of random subsets. 
In our experiments, we train the model with 10 randomly sampled subsets of 80 correspondences each.
At test time, we use 100 subsets with 20 correspondences each.
We evaluate the impact of those choices on performance and run time in \S~\ref{sec:exp_ablations}.

\subsection{Point Cloud Rendering} 
\label{sec:method_render}

The final step of our approach is to render the RGB-D images from the aligned point clouds. This provides us with our primary learning signals: photometric and depth consistency.
The core idea is that if the camera locations are estimated correctly, the point cloud renders will be consistent with the input images. 
We use differentiable rendering to project the colored point clouds onto an image using the estimated camera pose and known intrinsics. Our pipeline is very similar to Wiles \etal~\cite{wiles2019synsin}.

\begin{figure}[t]
\begin{center}
   \includegraphics[width=\linewidth]{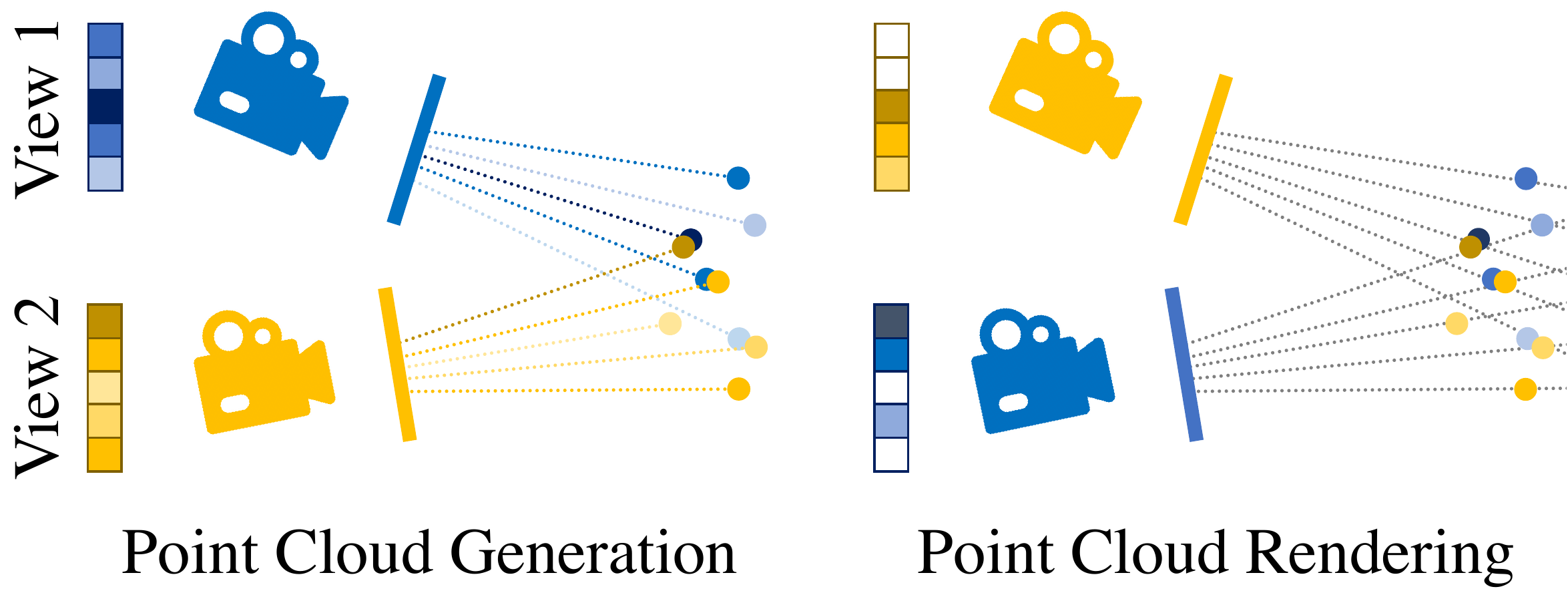}

\end{center}
   \caption{\textbf{Point Cloud Rendering. } 
   We project the views from both views, but only render from the alternative view; 
   \eg we render the points projected from view 2 in the perspective of view 1. 
   This can result in invalid pixels, visualized in white. 
    \textit{(For clarity, we show 1D projections in 2D space.)}
   }
\label{fig:mask_render}
\end{figure}

A naive approach of simply rendering both point clouds suffers from a degenerate solution: the rendering will be accurate even if the alignment is incorrect.
An extreme case of this would be to always estimate cameras looking in opposite directions.
In that case, each image is projected in a different location of space and the output will be consistent without alignment. 
We address this issue by forcing the network to render each view using only the other image's point cloud, as shown in Fig.~\ref{fig:mask_render}.  
This forces the network to learn consistent alignment as a correct reconstruction requires the mutually visible parts of the scene to be correctly aligned. 
This introduces another challenge: \textit{how to handle the non-mutually visible surfaces of the scene? }

While view synthesis approaches hallucinate the missing regions to output photo-realistic imagery~\cite{wiles2019synsin}, earlier work in differentiable SfM observed that the gradients coming from the hallucinated region negatively impact the learning~\cite{zhou2017unsupervised}. 
Our solution to this problem is to only evaluate the loss for valid pixels.
Valid pixels, as shown in Fig~\ref{fig:mask_render}, are ones for which rendering was possible; \ie, there were points along the viewing ray for those pixels. 
This is important in this work since invalid pixels can occur due to two reasons: non-mutually visible surfaces and pixels with missing depth.
While the first reason is due to our approach, the second reason for invalid pixels is governed by current depth sensors which do not produce a depth value for each pixel. 

In our experiments, we found that pose networks are very susceptible to the issues above; the network starts estimating very large poses within the first hundred iterations and never recovers. 
We also experimented with rendering the features and decoding them, similar to~\cite{wiles2019synsin}, but found that this resulted in worse alignment performance.

\subsection{Losses}

We use three consistency losses to train our model: photometric, depth, and correspondence.
The photometric and depth losses are the L1 losses applied between the rendered and input RGB-D frames. 
Those losses are masked to only apply to valid pixels, as discussed in \S~\ref{sec:method_render}. 
Additionally, we use the correspondence error calculated in Eq.~\ref{eq:corr_err} as our correspondence loss.
We weight the photometric and depth losses with a weighting of 1 while the correspondence loss receives a weighting of 0.1.

\begin{table*}[t]
\centering
\resizebox{\textwidth}{!}{

    \begin{tabular}{l cc cc  ccccc ccccc ccccc}
    \toprule
    &  & & \multicolumn{2}{c}{} & \multicolumn{5}{c}{Rotation} & \multicolumn{5}{c}{Translation} & \multicolumn{5}{c}{Chamfer}\\

    & & &  \multicolumn{2}{c}{Features}  
    & \multicolumn{3}{c}{Accuracy $\uparrow$} & \multicolumn{2}{c}{Error $\downarrow$} 
    & \multicolumn{3}{c}{Accuracy $\uparrow$} & \multicolumn{2}{c}{Error $\downarrow$} 
    & \multicolumn{3}{c}{Accuracy $\uparrow$} & \multicolumn{2}{c}{Error $\downarrow$} \\ 

    \cmidrule(lr){ 4-5}
    \cmidrule(lr){ 6- 8}   \cmidrule(lr){ 9-10}
    \cmidrule(lr){11-13}   \cmidrule(lr){14-15}
    \cmidrule(lr){16-18}   \cmidrule(lr){19-20}
  
    &  Train Set & Pose Sup. &  Visual & 3D 
    & $5^\circ$ & $10^\circ$ & $45^\circ$ & Mean & Med. 
    & 5 & 10 & 25 & Mean & Med. 
    & 1 & 5 & 10 & Mean & Med. 
    \\
    \midrule
    \multicolumn{5}{l}{\textbf{RANSAC + Feature Descriptors}}\\
    
    SIFT & - & \xmark & \cmark & \xmark &
    55.2 & 75.7 & 89.2 & 18.6 &  4.3 & 17.7 & 44.5 & 79.8 & 26.5 & 11.2 & 38.1 & 70.6 & 78.3 & 42.6 &  1.7
    \\

    SuperPoint~\cite{detone2018superpoint} & - & \xmark & \cmark & \xmark &
    65.5 & 86.9 & 96.6 &  8.9 &  3.6 & 21.2 & 51.7 & 88.0 & 16.1 &  9.7 & 45.7 & 81.1 & 88.2 & 19.2 &  1.2 
    \\

    FCGF~\cite{choy2019fully} & - & \xmark & \xmark & \cmark &
    70.2 & 87.7 & 96.2 &  9.5 &  3.3 & 27.5 & 58.3 & 82.9 & 23.6 &  8.3 & 52.0 & 78.0 & 83.7 & 24.4 &  0.9  
    \\

    \midrule
    \multicolumn{5}{l}{\textbf{Supervised Geometric Approaches}}\\
    
    DGR~\cite{choy2020deep} & 3D Match & \cmark & \xmark & \cmark &
    81.1 & 89.3 & 94.8 &  9.4 &  1.8 & 54.5 & 76.2 & 88.7 & 18.4 &  4.5 & 70.5 & 85.5 & 89.0 & 13.7 &  0.4
    \\
    
    3D MV Reg~\cite{gojcic2020learning} & 3D Match & \cmark & \xmark & \cmark &
    87.7 & 93.2 & 97.0 &  6.0 &  1.2 & 69.0 & 83.1 & 91.8 & 11.7 &  2.9 & 78.9 & 89.2 & 91.8 & 10.2 &  0.2 
    \\

    \midrule
    
    Ours & 3D Match & \xmark & \cmark & \xmark &    
    87.6 & 93.1 & 98.3 &  4.3 &  1.0 & 69.2 & 84.0 & 93.8 &  9.5 &  2.8 & 79.7 & 91.3 & 94.0 &  7.2 &  0.2 
    \\

    Ours & ScanNet & \xmark & \cmark & \xmark & 
    92.7 & 95.8 & 98.5 &  3.4 &  0.8 & 77.2 & 89.6 & 96.1 &  7.3 &  2.3 & 86.0 & 94.6 & 96.1 &  5.9 &  0.1 \\

    \bottomrule
    \end{tabular}
}
\label{tab:pose_scannet}
\caption{\textbf{Pairwise Registration on ScanNet.}
We outperform existing registration pipelines that use traditional and learned, visual and geometric feature descriptors with a RANSAC estimator. 
Furthermore, we perform on-par with supervised geometric matching methods that were trained on 3D Match, demonstrating the utility of unsupervised training in this domain.
\textit{Pose Sup.} indicates pose supervision.
}

\end{table*}

\section{Experiments}
\label{sec:experiments}
We now empirically evaluate our model on pairwise point cloud registration.
Our experiments aim to answer several questions: 
(1) does unsupervised training provide us with useful features for alignment?; 
(2) can RGB-D video alleviate the need for the pose supervision required by geometric registration approaches?; 
(3) how do the different components of the model contribute to its performance? 

We address those questions by evaluating our approach on two datasets of indoor scenes: ScanNet~\cite{dai2017scannet} and 3DMatch~\cite{zeng20163dmatch}.
We find that our approach achieves better registration accuracy than off-the-shelf visual and geometric feature descriptors (\S~\ref{sec:exp_pcreg}).
We also find that our approach performs on-par with supervised geometric registration approaches despite using significantly simpler correspondence matching and alignment algorithms; supporting our claim that RGB-D video can alleviate the need for pose supervision.
Finally, we analyze our model components through several key ablations (\S~\ref{sec:exp_ablations}). 

\lsparagraph{Datasets.} 
We evaluate our approach using ScanNet~\cite{dai2017scannet} and 3D Match~\cite{zeng20163dmatch}. 
ScanNet contains RGB-D images and ground-truth camera poses for 1513 scenes, while 3D Match is a much smaller dataset with a total of 101 scenes.
We use the official data split of 1045/156/312 scenes for train/val/test for ScanNet.
3D Match only provides a train/test split, so we further divide the train split into train and validation; resulting in 71/11/19 RGB-D sequences for train/val/test split.
We generate view pairs by sampling image pairs that are 20 frames apart.
We sample the training scenes more densely by sampling all pairs that are 20 frames apart. 
This results in 1594k/12.6k/26k ScanNet pairs and 122k/1.5k/1.5k 3D Match pairs.

\lsparagraph{Baselines. }
We compare our model to several learned and non-learned point cloud registration approaches. 
Since we are interested in the unsupervised setting, we first compare against methods that do not require pose supervision. 
Our first set of baselines use off-the-shelf keypoint detectors and descriptors with RANSAC~\cite{RANSAC} as the robust estimator. For all these baselines, we use Open3D's RANSAC implementation~\cite{open3d}.
Despite being proposed over a decade ago, SIFT features are still used and serve as a strong baseline for a non-learned method. 
SuperPoint~\cite{detone2018superpoint} is a recently proposed approach for keypoint detection and description and has achieved state of the art performance in correspondence matching on several benchmarks.
Finally, FCGF~\cite{choy2019fully} is a recently proposed geometric feature descriptor that has also achieved state-of-the-art performance on several 3D correspondence benchmarks. 
Furthermore, FCGF features have been used by several recent approaches for point cloud registration without further fine-tuning~\cite{choy2020deep,gojcic2020learning}. 

We also compare against two supervised geometric registration approaches: DGR~\cite{choy2020deep} and 3D MV Registration~\cite{gojcic2020learning}. 
Both of these approaches operate on FCGF point cloud embeddings as their input and learn how to extract good correspondences between pairs. 
There are two salient differences between our approaches: First, our approach is unsupervised, while those approaches rely on pose supervision. Second, our approach operates on RGB-D, while those approaches use the FCGF embeddings of the point cloud without relying on the images. This comparison demonstrates how leveraging the currently ignored RGB modality could alleviate the need for pose supervision and pretrained descriptors. We emphasize that we use the weights provided by the authors which were trained on the 3D Match Geometric Registration benchmark.

\lsparagraph{Training Details. }
We train our models with the Adam~\cite{kingma2015adam} optimizer with a learning rate of $10^{-4}$ and momentum parameters of (0.9, 0.99). 
We train each model for 200K iterations. We implement our models in PyTorch~\cite{pytorch3d}, while making extensive use of PyTorch3D~\cite{pytorch3d} and Open3D~\cite{open3d}. 

\begin{table*}[t]
\centering
\resizebox{\textwidth}{!}{

    \begin{tabular}{l c cc  ccccc ccccc ccccc}
    \toprule
    & &
    & \multicolumn{5}{c}{Rotation} & \multicolumn{5}{c}{Translation} & \multicolumn{5}{c}{Chamfer}\\

    & \multicolumn{2}{c}{Ablation} 
    & \multicolumn{3}{c}{Accuracy $\uparrow$} & \multicolumn{2}{c}{Error $\downarrow$} 
    & \multicolumn{3}{c}{Accuracy $\uparrow$} & \multicolumn{2}{c}{Error $\downarrow$} 
    & \multicolumn{3}{c}{Accuracy $\uparrow$} & \multicolumn{2}{c}{Error $\downarrow$} \\ 
    
    \cmidrule(lr){ 2- 3}
    \cmidrule(lr){ 4- 6}   \cmidrule(lr){ 7- 8}
    \cmidrule(lr){ 9-11}   \cmidrule(lr){12-13}
    \cmidrule(lr){14-16}   \cmidrule(lr){17-18}
  
    & Train & Test
    & $5^\circ$ & $10^\circ$ & $45^\circ$ & Mean & Med. 
    & 5 & 10 & 25 & Mean & Med. 
    & 1 & 5 & 10 & Mean & Med.
    \\
    \midrule

Full Model &  &  &
92.7 & 95.8 & 98.5 &  3.4 &  0.8 & 77.2 & 89.6 & 96.1 &  7.3 &  2.3 & 86.0 & 94.6 & 96.1 &  5.9 &  0.1 \\

Model with Joint Rendering & \cmark &  &
84.7 & 91.1 & 97.5 &  5.5 &  1.1 & 65.3 & 81.2 & 92.0 & 12.0 &  3.1 & 76.5 & 89.1 & 92.4 &  9.0 &  0.2 \\

\midrule

\quad - Randomized Optimization \hspace{0.5cm}  & \cmark & \cmark &
86.0 & 93.0 & 98.1 &  4.7 &  1.3 & 59.3 & 79.6 & 93.1 & 10.8 &  3.9 & 73.5 & 90.0 & 93.4 &  8.2 &  0.3 \\
\quad - Ratio Test  & \cmark & \cmark &
77.6 & 88.9 & 97.1 &  6.9 &  1.9 & 48.5 & 70.4 & 89.2 & 15.1 &  5.2 & 64.1 & 84.9 & 90.0 & 10.5 &  0.5 \\
\midrule

\quad - Randomized Optimization & \cmark & \xmark &
94.6 & 97.0 & 98.9 &  2.8 &  0.8 & 80.3 & 91.8 & 97.1 &  6.1 &  2.1 & 88.5 & 95.9 & 97.2 &  5.0 &  0.1
\\
\quad - Ratio Test  & \cmark & \xmark &
86.0 & 92.8 & 98.7 &  4.1 &  1.1 & 64.8 & 82.0 & 93.6 &  9.3 &  3.2 & 76.7 & 90.5 & 93.8 &  6.8 &  0.2
\\
\midrule

\quad - Randomized Optimization & \xmark & \cmark &
81.1 & 90.4 & 97.6 &  5.8 &  1.6 & 52.3 & 74.1 & 90.6 & 13.3 &  4.7 & 67.6 & 86.8 & 91.2 &  9.7 &  0.4  \\
\quad - Ratio Test  & \xmark & \cmark &
47.2 & 67.4 & 91.8 & 16.6 &  5.5 & 20.6 & 40.4 & 69.3 & 35.5 & 13.4 & 32.4 & 60.3 & 71.2 & 27.8 &  2.8  \\
\bottomrule
\end{tabular}
}
\label{tab:ablations}
\caption{\textbf{Ablation Results.} 
Our ablation experiments demonstrate the utility of the ratio test for correspondence filtering. Furthermore, we find that some ablations can improve model performance when used for training, but not for testing. 
}
\end{table*}

\subsection{Pairwise Registration}
\label{sec:exp_pcreg}

We first evaluate our approach on point cloud registration. 
Given two RGB-D images, we estimate the 6-DOF pose that would best align the first input image to the second. 
The transformation is represented by a rotation matrix $\mathbf{R}$ and translation vector $\mathbf{t}$.

\lsparagraph{Evaluation Metrics.}
We evaluate pairwise registration by evaluating the pose prediction as well as the chamfer distance between the estimated and ground-truth alignments.
We compute the angular and translation errors as follows:
$$
E_{\text{rotation}} = \arccos(\frac{Tr(\mathbf{R}_{pr}\mathbf{R}_{gt}^\top) - 1}{2}), 
$$$$
E_{\text{translation}} = ||\mathbf{t}_{pr} - \mathbf{t}_{gt}||_2 .
$$
We report the translation error in centimeters and the rotation errors in degrees.

While pose gives us a good measure of performance, some scenes are inherently ambiguous and multiple alignments can explain the scene appearance; \eg, walls, floors, symmetric objects. 
To address these cases, we compute the chamfer distance between the scene and our reconstruction. 
Given two point clouds where $\mathcal{P}$ represents the correct alignment of the scene and $\mathcal{Q}$ represents our reconstruction of the scene, 
we can define the closest pairs between the point clouds as set $\Lambda_{\mathcal{P}, \mathcal{Q}} = \{(p, \argmin_{q \in \mathcal{Q}} ||p - q||) : p \in \mathcal{P})$.
We then compute the chamfer error as follows:
$$
E_{\text{cham}} = 
|\mathcal{P}|^{-1} \sum_{\mathclap{({p, q) \in \Lambda_{\mathcal{P}, \mathcal{Q}}}}} ||\mathbf{x}_p - \mathbf{x}_q|| 
+ 
|\mathcal{Q}|^{-1}  \sum_{\mathclap{{(q, p) \in \Lambda_{\mathcal{Q}, \mathcal{P}}}}} ||\mathbf{x}_q - \mathbf{x}_p||.
$$

For each of these error metrics, we report the mean and median errors over the dataset as well as the accuracy for different thresholds.

We conduct our experiments on ScanNet and report the results in Table~\ref{tab:pose_scannet}. 
We find that our model learns accurate point cloud registration; outperforming prior feature descriptors and performing on-par with supervised geometric registration approaches.
We next analyze our results through the questions posed at the start of this section. 

\lsparagraph{Does unsupervised learning improve over off-the-shelf descriptors? }
Yes. We evaluate our approach against the traditional pipeline for registration: feature extraction using an off-the-shelf keypoint descriptor and alignment via RANSAC. 
We show large performance gains over both traditional and learned descriptors.
It is important to note that FCGF and SuperPoint currently represent the state-of-the-art for feature descriptors.
Furthermore, both methods have been used directly, without further fine-tuning, to achieve the highest performance on image registration benchmarks~\cite{sarlin2020superglue} and geometric registration benchmarks~\cite{choy2020deep,gojcic2020learning}.
We also find that our approach learns features that can generalize to similar datasets. As shown in Table~\ref{tab:pose_scannet}, our model trained on 3D Match outperforms the off-the-shelf descriptors while being competitive with supervised geometric registration approaches.

\lsparagraph{Does RGB-D training alleviate the need for pose supervision? }
Yes. 
We compare our approach to two recently proposed supervised point cloud registration approaches: DGR~\cite{choy2020deep} and 3D Multi-view Registration~\cite{gojcic2020learning}. 
Since their model was trained on 3D Match, we also train our model on 3D match and report the numbers. 
We find that our model is competitive with supervised approaches when trained on their dataset, and can outperform them when trained on ScanNet. 
However, a direct comparison is more nuanced since those two classes of methods differ in two key ways: training supervision and input modality.

We argue that the recent rise in RGB-D cameras on both hand-held devices and robotic systems supports our setup. 
First, the rise in devices suggests a corresponding increase in RGB-D raw data that will not necessarily be annotated with pose information. 
This increase provides a great opportunity for unsupervised learning to leverage this data stream. 
Second, while there are cases where depth sensing might be the better or only option (\eg, dark environment or highly reflective surfaces.), there are many cases where one has access to both RGB and depth information. 
The ability to leverage both can increase the effectiveness and robustness of a registration system. 
Finally, while we only learn visual features in this work, we note that our approach is easily extensible to learning both geometric and visual features since it is agnostic to how the features are calculated.

\subsection{Ablations}
\label{sec:exp_ablations}

We perform several ablation studies to better understand the model’s performance and its various components.
In particular, we are interested in better understanding the impact of the optimization and rendering parameters on the overall model performance. 
While some ablations can only be applied during training (\eg, rendering choice), ablations that affect the correspondence estimation and fitting  can be selectively applied during training, inference, or both. 
Hence, we consider all the variants.

\vspace{0.2cm}
\lsparagraph{Joint Rendering.}
Our first ablation investigates the impact of our rendering choices by rendering the output images from the joint point cloud. 
In \S~\ref{sec:method_render}, we discuss rendering alternate views to force the model to align the pointclouds to produce accurate renders. 
As shown in Table~\ref{tab:ablations}, we find that naively rendering the joint point cloud results in a significant performance drop.
This supports our claim that a joint render would negatively impact the features learned since the model can achieve good photometric consistency even if the pointclouds are not accurately aligned.

\vspace{0.2cm}
\lsparagraph{Ratio Test.}
In our approach, we use Lowe's ratio test to estimate the weight for each correspondence. 
We ablate this component by instead using the feature distance between the corresponding points to rank the correspondences.
Since this ablation can be applied to training or inference independently, we apply it to training, inference, or both. 
Our results indicate that the ratio test is critical to our model's performance, as ablating it results in the largest performance drop.
This supports our initial claims about the utility of the ratio test as a strong heuristic for filtering correspondences. 
It is worth noting that Lowe's ratio test~\cite{lowe2004distinctive} shows incredible efficacy in determining correspondence weights; a function often undertaken by far more complex models in recent work~\cite{choy2020deep,gojcic2020learning,ranftl2018deepfundamental,sarlin2020superglue}.
Our approach is able to perform well using such a simple filtering heuristic since it is also learning the features, not just matching them.

\vspace{0.2cm}
\lsparagraph{Randomized Subsets.}
In our model, we estimate $t$ transformations based on $t$ randomly sampled subsets. This is inspired by RANSAC~\cite{RANSAC} as it allows us to better handle outliers. 
We ablate this module by estimating a single transformation based on all the correspondences. 
Similar to the ratio test, this ablation can be applied to training or inference independently.
As shown in Table~\ref{tab:ablations}, ablating this component at test time results in a significant drop in performance. 
Interestingly, we find that applying it during training and relieving it during testing improves performance.
We posit that this ablation acts similarly to DropOut~\cite{srivastava2014dropout} which forces the model to predict using a subset of the features and is only applied during training. As a result, the model is forced to learn better features during training, while gaining the benefits of randomized optimization during inference. 

\vspace{0.2cm}
\lsparagraph{Number of subsets. } 
We find that the number of subsets chosen has a significant impact on both run-time and performance.
During training, we sample 10 subsets of 80 correspondences each. During testing, we sample 100 subsets of 80 correspondences each. 
For this set of experiments, we used the same pretrained weights and only vary the number of subsets used. Each subset still contains 80 correspondences. 
As shown in Table~\ref{tab:runtime}, using a larger number of subsets improves the performance while also increasing the run-time. 
Additionally, we find that the performance gains saturate at 100 subsets.

\section{Conclusion}
\label{sec:conclusion}
We present an unsupervised, end-to-end approach to pairwise RGB-D point cloud registration. 
We observe that existing approaches to point cloud registration rely on pose supervision for learning geometric point cloud alignment.
However, with the increase in cameras with depth sensors, we expect a large stream of unannotated RGB-D data. 
This provides us with an opportunity to leverage unsupervised learning for more robust RGB-D point cloud registration. 

To this end, we propose using view synthesis as a task for unsupervised point cloud registration via differentiable alignment and rendering.
At the core of our approach is the notion of achieving geometric alignment through training a model on photometric consistency. 
Our approach learns to extract features from RGB-D data that allow it to both register and render the input frames. 
We show that our approach outperforms current state-of-the-art feature descriptors with RANSAC as well as supervised geometric registration approaches. This supports our initial premise of using RGB-D data to alleviate the need for pose supervision. 

While our implementation relies solely on features extracted from RGB, our approach does not necessitate this. 
Specifically, our approach could be extended to learning geometric features for correspondence estimation. 
Furthermore, while we find that the ratio test allows us to achieve highly accurate registration, it would be interesting to explore whether recently proposed supervised correspondence filtering algorithms can be adapted for unsupervised training as well as how they would compare to the simple ratio test heuristic. 

\begin{table}[t]
\resizebox{\columnwidth}{!}{
    \begin{tabular}{r cc cc  cc r}
    \toprule
    & \multicolumn{2}{c}{Rotation} 
    & \multicolumn{2}{c}{Translation} 
    & \multicolumn{2}{c}{Chamfer} 
    &
    \\
    
    \cmidrule(lr){ 2- 3}
    \cmidrule(lr){ 4- 5}
    \cmidrule(lr){ 6- 7}
    
    Subsets  
    & Mean & Med.
    & Mean & Med.
    & Mean & Med. &
    Time (ms)
    \\
    \midrule
    
    5   & 4.8 &  1.2 & 10.5 &  3.4 &  7.8 &  0.2 &  50.4 $\pm$~0.3   \\
    10  & 4.2 &  1.0 &  9.2 &  2.9 &  7.1 &  0.2 &  60.2 $\pm$~0.3   \\
    20  & 3.8 &  0.9 &  8.4 &  2.6 &  6.7 &  0.2 &  79.2 $\pm$~0.5  \\
    50  & 3.5 &  0.9 &  7.7 &  2.4 &  6.0 &  0.1 & 135.4 $\pm$~1.1  \\
    100 & 3.4 &  0.8 &  7.3 &  2.3 &  5.9 &  0.1 & 239.6 $\pm$~1.2  \\
    200 & 3.3 &  0.8 &  7.2 &  2.2 &  5.9 &  0.1 & 425.2 $\pm$~6.1  \\

    \bottomrule
\end{tabular}
}
\label{tab:runtime}
\caption{\textbf{Run-time Analysis.} 
We find that using a larger number of random subsets improves our performance while also increasing the inference time.
This trade-off between performance and run-time could be used to tune the model based on the use case. 
}
\end{table}

\vspace{6pt} \noindent
\textbf{Acknowledgments}
We would like to thank the anonymous reviewers for their valuable comments and suggestions. 
We also thank Nilesh Kulkarni, Karan Desai, Richard Higgins, and Max Smith for many helpful discussions and feedback on early drafts of this work.

\clearpage
\newpage

{\small
\bibliographystyle{format/ieee}
\bibliography{bibliography}

\begin{thebibliography}{10}\itemsep=-1pt

\bibitem{aoki2019pointnetlk}
Yasuhiro Aoki, Hunter Goforth, Rangaprasad~Arun Srivatsan, and Simon Lucey.
\newblock Pointnetlk: Robust \& efficient point cloud registration using
  pointnet.
\newblock In {\em CVPR}, 2019.

\bibitem{arun1987least}
KS Arun, TS Huang, and SD Blostein.
\newblock Least square fitting of two 3-d point sets.
\newblock In {\em TPAMI}, 1987.

\bibitem{bai2020d3feat}
Xuyang Bai, Zixin Luo, Lei Zhou, Hongbo Fu, Long Quan, and Chiew-Lan Tai.
\newblock D3feat: Joint learning of dense detection and description of 3d local
  features.
\newblock In {\em CVPR}, 2020.

\bibitem{bay2006surf}
Herbert Bay, Tinne Tuytelaars, and Luc Van~Gool.
\newblock Surf: Speeded up robust features.
\newblock In {\em ECCV}, 2006.

\bibitem{bian2020depth}
Jia-Wang Bian, Huangying Zhan, Naiyan Wang, Tat-Jun Chin, Chunhua Shen, and Ian
  Reid.
\newblock Unsupervised depth learning in challenging indoor video: Weak
  rectification to rescue.
\newblock {\em arXiv}, 2020.

\bibitem{brachmann2017dsac}
Eric Brachmann, Alexander Krull, Sebastian Nowozin, Jamie Shotton, Frank
  Michel, Stefan Gumhold, and Carsten Rother.
\newblock Dsac-differentiable ransac for camera localization.
\newblock In {\em CVPR}, 2017.

\bibitem{brachmann2019neural}
Eric Brachmann and Carsten Rother.
\newblock Neural-guided ransac: Learning where to sample model hypotheses.
\newblock In {\em ICCV}, 2019.

\bibitem{casser2019depthprediction}
Vincent~Michael Casser, Soeren Pirk, Reza Mahjourian, and Anelia Angelova.
\newblock Depth prediction without the sensors: Leveraging structure for
  unsupervised learning from monocular videos.
\newblock In {\em AAAI}, 2019.

\bibitem{choy2020deep}
Christopher Choy, Wei Dong, and Vladlen Koltun.
\newblock Deep global registration.
\newblock In {\em CVPR}, 2020.

\bibitem{choy2019fully}
Christopher Choy, Jaesik Park, and Vladlen Koltun.
\newblock Fully convolutional geometric features.
\newblock In {\em ICCV}, 2019.

\bibitem{choy20163d}
Christopher~B Choy, Danfei Xu, JunYoung Gwak, Kevin Chen, and Silvio Savarese.
\newblock 3d-r2n2: A unified approach for single and multi-view 3d object
  reconstruction.
\newblock In {\em ECCV}, 2016.

\bibitem{dai2017scannet}
Angela Dai, Angel~X. Chang, Manolis Savva, Maciej Halber, Thomas Funkhouser,
  and Matthias Nie{\ss}ner.
\newblock Scannet: Richly-annotated 3d reconstructions of indoor scenes.
\newblock In {\em CVPR}, 2017.

\bibitem{deng2018ppf}
Haowen Deng, Tolga Birdal, and Slobodan Ilic.
\newblock Ppf-foldnet: Unsupervised learning of rotation invariant 3d local
  descriptors.
\newblock In {\em ECCV}, 2018.

\bibitem{deng20193d}
Haowen Deng, Tolga Birdal, and Slobodan Ilic.
\newblock 3d local features for direct pairwise registration.
\newblock In {\em CVPR}, 2019.

\bibitem{desai2020virtex}
Karan Desai and Justin Johnson.
\newblock Virtex: Learning visual representations from textual annotations.
\newblock {\em arXiv}, 2020.

\bibitem{detone2018superpoint}
Daniel DeTone, Tomasz Malisiewicz, and Andrew Rabinovich.
\newblock Superpoint: Self-supervised interest point detection and description.
\newblock In {\em CVPR Workshops}, 2018.

\bibitem{doersch2015unsupervised}
Carl Doersch, Abhinav Gupta, and Alexei~A Efros.
\newblock Unsupervised visual representation learning by context prediction.
\newblock In {\em ICCV}, 2015.

\bibitem{dusmanu2019d2}
Mihai Dusmanu, Ignacio Rocco, Tomas Pajdla, Marc Pollefeys, Josef Sivic,
  Akihiko Torii, and Torsten Sattler.
\newblock D2-net: A trainable cnn for joint detection and description of local
  features.
\newblock In {\em CVPR}, 2019.

\bibitem{elbanani2020novelviewpoints}
Mohamed El~Banani, Jason~J. Corso, and David Fouhey.
\newblock Novel object viewpoint estimation through reconstruction alignment.
\newblock In {\em CVPR}, 2020.

\bibitem{RANSAC}
Martin~A. Fischler and Robert~C. Bolles.
\newblock Random sample consensus: A paradigm for model fitting with
  applications to image analysis and automated cartography.
\newblock {\em Commun. ACM}, 24(6), 1981.

\bibitem{gidaris2018unsupervised}
Spyros Gidaris, Praveer Singh, and Nikos Komodakis.
\newblock Unsupervised representation learning by predicting image rotations.
\newblock In {\em ICLR}, 2018.

\bibitem{gojcic2020learning}
Zan Gojcic, Caifa Zhou, Jan~D. Wegner, Leonidas~J. Guibas, and Tolga Birdal.
\newblock Learning multiview 3d point cloud registration.
\newblock In {\em CVPR}, 2020.

\bibitem{gojcic2019perfect}
Zan Gojcic, Caifa Zhou, Jan~D Wegner, and Andreas Wieser.
\newblock The perfect match: 3d point cloud matching with smoothed densities.
\newblock In {\em CVPR}, 2019.

\bibitem{gower1975generalized}
John~C Gower.
\newblock Generalized procrustes analysis.
\newblock {\em Psychometrika}, 40(1):33--51, 1975.

\bibitem{goyal2019scaling}
Priya Goyal, Dhruv Mahajan, Abhinav Gupta, and Ishan Misra.
\newblock Scaling and benchmarking self-supervised visual representation
  learning.
\newblock In {\em ICCV}, 2019.

\bibitem{han2015matchnet}
Xufeng Han, Thomas Leung, Yangqing Jia, Rahul Sukthankar, and Alexander~C Berg.
\newblock Matchnet: Unifying feature and metric learning for patch-based
  matching.
\newblock In {\em CVPR}, 2015.

\bibitem{hertz2020pointgmm}
Amir Hertz, Rana Hanocka, Raja Giryes, and Daniel Cohen-Or.
\newblock Pointgmm: A neural gmm network for point clouds.
\newblock In {\em CVPR}, 2020.

\bibitem{huang2020featureregistration}
Xiaoshui Huang, Guofeng Mei, and Jian Zhang.
\newblock Feature-metric registration: A fast semi-supervised approach for
  robust point cloud registration without correspondences.
\newblock In {\em CVPR}, 2020.

\bibitem{insafutdinov2018pointclouds}
Eldar Insafutdinov and Alexey Dosovitskiy.
\newblock Unsupervised learning of shape and pose with differentiable point
  clouds.
\newblock In {\em NeurIPS}, 2018.

\bibitem{johnson1999using}
Andrew~E. Johnson and Martial Hebert.
\newblock Using spin images for efficient object recognition in cluttered 3d
  scenes.
\newblock {\em TPAMI}, 1999.

\bibitem{kabsch1976solution}
Wolfgang Kabsch.
\newblock A solution for the best rotation to relate two sets of vectors.
\newblock {\em Acta Crystallographica Section A: Crystal Physics, Diffraction,
  Theoretical and General Crystallography}, 32(5):922--923, 1976.

\bibitem{kalantari2016learning}
Nima~Khademi Kalantari, Ting-Chun Wang, and Ravi Ramamoorthi.
\newblock Learning-based view synthesis for light field cameras.
\newblock {\em ACM Transactions on Graphics (TOG)}, 35(6):1--10, 2016.

\bibitem{kar2017learning}
Abhishek Kar, Christian H{\"a}ne, and Jitendra Malik.
\newblock Learning a multi-view stereo machine.
\newblock In {\em NeurIPS}, 2017.

\bibitem{kingma2015adam}
Diederik Kingma and Jimmy Ba.
\newblock Adam: A method for stochastic optimization.
\newblock In {\em ICLR}, 2015.

\bibitem{kulkarni2020acsm}
Nilesh Kulkarni, Abhinav Gupta, David~F Fouhey, and Shubham Tulsiani.
\newblock Articulation-aware canonical surface mapping.
\newblock In {\em CVPR}, 2020.

\bibitem{Li2020e2e3ddescriptors}
Lei Li, Siyu Zhu, Hongbo Fu, Ping Tan, and Chiew-Lan Tai.
\newblock End-to-end learning local multi-view descriptors for 3d point clouds.
\newblock In {\em The IEEE Conference on Computer Vision and Pattern
  Recognition (CVPR)}, 2020.

\bibitem{longuet1981computer}
Hugh~Christopher Longuet-Higgins.
\newblock A computer algorithm for reconstructing a scene from two projections.
\newblock {\em Nature}, 1981.

\bibitem{lowe2004distinctive}
David~G Lowe.
\newblock Distinctive image features from scale-invariant keypoints.
\newblock {\em IJCV}, 2004.

\bibitem{lu2019deepvcp}
Weixin Lu, Guowei Wan, Yao Zhou, Xiangyu Fu, Pengfei Yuan, and Shiyu Song.
\newblock Deepvcp: An end-to-end deep neural network for point cloud
  registration.
\newblock In {\em ICCV}, 2019.

\bibitem{Mahjourian2018depthicp}
Reza Mahjourian, Martin Wicke, and Anelia Angelova.
\newblock Unsupervised learning of depth and egomotion from monocular video
  using 3d geometric constraints.
\newblock In {\em CVPR}, 2018.

\bibitem{martin2020nerf}
Ricardo Martin-Brualla, Noha Radwan, Mehdi~SM Sajjadi, Jonathan~T Barron,
  Alexey Dosovitskiy, and Daniel Duckworth.
\newblock Nerf in the wild: Neural radiance fields for unconstrained photo
  collections.
\newblock {\em arXiv}, 2020.

\bibitem{mildenhall2020nerf}
Ben Mildenhall, Pratul~P Srinivasan, Matthew Tancik, Jonathan~T Barron, Ravi
  Ramamoorthi, and Ren Ng.
\newblock Nerf: Representing scenes as neural radiance fields for view
  synthesis.
\newblock {\em ECCV}, 2020.

\bibitem{moravec1981rover}
Hans~P. Moravec.
\newblock Rover visual obstacle avoidance.
\newblock In {\em IJCAI}, 1981.

\bibitem{niklaus20193d}
Simon Niklaus, Long Mai, Jimei Yang, and Feng Liu.
\newblock 3d ken burns effect from a single image.
\newblock {\em ACM Transactions on Graphics (TOG)}, 38(6):1--15, 2019.

\bibitem{penner2017soft}
Eric Penner and Li Zhang.
\newblock Soft 3d reconstruction for view synthesis.
\newblock {\em ACM Transactions on Graphics (TOG)}, 36(6):1--11, 2017.

\bibitem{pomerleau2015review}
Fran{\c{c}}ois Pomerleau, Francis Colas, and Roland Siegwart.
\newblock A review of point cloud registration algorithms for mobile robotics.
\newblock {\em Foundations and Trends in Robotics}, 4(1):1--104, 2015.

\bibitem{qian2020associative3d}
Shengyi Qian, Linyi Jin, and David~F. Fouhey.
\newblock Associative3d: Volumetric reconstruction from sparse views.
\newblock In {\em ECCV}, 2020.

\bibitem{ranftl2018deepfundamental}
Rene Ranftl and Vladlen Koltun.
\newblock Deep fundamental matrix estimation.
\newblock In {\em ECCV}, 2018.

\bibitem{pytorch3d}
Nikhila Ravi, Jeremy Reizenstein, David Novotny, Taylor Gordon, Wan-Yen Lo,
  Justin Johnson, and Georgia Gkioxari.
\newblock Accelerating 3d deep learning with pytorch3d.
\newblock {\em arXiv}, 2020.

\bibitem{rublee2011orb}
Ethan Rublee, Vincent Rabaud, Kurt Konolige, and Gary Bradski.
\newblock Orb: An efficient alternative to sift or surf.
\newblock In {\em ICCV}, 2011.

\bibitem{rusu2009fast}
Radu~Bogdan Rusu, Nico Blodow, and Michael Beetz.
\newblock Fast point feature histograms (fpfh) for 3d registration.
\newblock In {\em ICRA}, 2009.

\bibitem{sarlin2020superglue}
Paul-Edouard Sarlin, Daniel DeTone, Tomasz Malisiewicz, and Andrew Rabinovich.
\newblock Superglue: Learning feature matching with graph neural networks.
\newblock In {\em CVPR}, 2020.

\bibitem{schonberger2016structure}
Johannes~L Schonberger and Jan-Michael Frahm.
\newblock Structure-from-motion revisited.
\newblock In {\em CVPR}, 2016.

\bibitem{shin20193d}
Daeyun Shin, Zhile Ren, Erik~B Sudderth, and Charless~C Fowlkes.
\newblock 3d scene reconstruction with multi-layer depth and epipolar
  transformers.
\newblock In {\em ICCV}, 2019.

\bibitem{srinivasan2019pushing}
Pratul~P Srinivasan, Richard Tucker, Jonathan~T Barron, Ravi Ramamoorthi, Ren
  Ng, and Noah Snavely.
\newblock Pushing the boundaries of view extrapolation with multiplane images.
\newblock In {\em CVPR}, 2019.

\bibitem{srinivasan2017learning}
Pratul~P Srinivasan, Tongzhou Wang, Ashwin Sreelal, Ravi Ramamoorthi, and Ren
  Ng.
\newblock Learning to synthesize a 4d rgbd light field from a single image.
\newblock In {\em ICCV}, 2017.

\bibitem{srivastava2014dropout}
Nitish Srivastava, Geoffrey Hinton, Alex Krizhevsky, Ilya Sutskever, and Ruslan
  Salakhutdinov.
\newblock Dropout: A simple way to prevent neural networks from overfitting.
\newblock {\em JMLR}, 2014.

\bibitem{tang2018banet}
Chengzhou Tang and Ping Tan.
\newblock Ba-net: Dense bundle adjustment networks.
\newblock In {\em ICLR}, 2018.

\bibitem{teed2019deepv2d}
Zachary Teed and Jia Deng.
\newblock Deepv2d: Video to depth with differentiable structure from motion.
\newblock In {\em ICLR}, 2019.

\bibitem{tian2019contrastive}
Yonglong Tian, Dilip Krishnan, and Phillip Isola.
\newblock Contrastive multiview coding.
\newblock {\em arXiv}, 2019.

\bibitem{torr1997development}
Philip~HS Torr and David~William Murray.
\newblock The development and comparison of robust methods for estimating the
  fundamental matrix.
\newblock {\em IJCV}, 1997.

\bibitem{tulsiani2018multiviewconsistency}
Shubham Tulsiani, Alexei~A Efros, and Jitendra Malik.
\newblock Multi-view consistency as supervisory signal for learning shape and
  pose prediction.
\newblock In {\em CVPR}, 2018.

\bibitem{tulsiani2015viewpoints}
Shubham Tulsiani and Jitendra Malik.
\newblock Viewpoints and keypoints.
\newblock In {\em CVPR}, 2015.

\bibitem{ummenhofer2017demon}
Benjamin Ummenhofer, Huizhong Zhou, Jonas Uhrig, Nikolaus Mayer, Eddy Ilg,
  Alexey Dosovitskiy, and Thomas Brox.
\newblock Demon: Depth and motion network for learning monocular stereo.
\newblock In {\em CVPR}, 2017.

\bibitem{vijayanarasimhan17sfmnet}
Sudheendra Vijayanarasimhan, Susanna Ricco, Cordelia Schmid, Rahul Sukthankar,
  and Katerina Fragkiadaki.
\newblock Sfm-net: Learning of structure and motion from video.
\newblock In {\em ArXiv}, 2017.

\bibitem{wang2019prnet}
Yue Wang and Justin Solomon.
\newblock Prnet: Self-supervised learning for partial-to-partial registration.
\newblock {\em NeurIPS}, 2019.

\bibitem{wang2019dcp}
Yue Wang and Justin~M Solomon.
\newblock Deep closest point: Learning representations for point cloud
  registration.
\newblock In {\em ICCV}, 2019.

\bibitem{wiles2019synsin}
Olivia Wiles, Georgia Gkioxari, Richard Szeliski, and Justin Johnson.
\newblock Synsin: End-to-end view synthesis from a single image.
\newblock In {\em CVPR}, 2020.

\bibitem{yew20183dfeatnet}
Zi~Jian Yew and Gim~Hee Lee.
\newblock 3dfeat-net: Weakly supervised local 3d features for point cloud
  registration.
\newblock In {\em ECCV}, 2018.

\bibitem{yew2020RPMNet}
Zi~Jian Yew and Gim~Hee Lee.
\newblock Rpm-net: Robust point matching using learned features.
\newblock In {\em CVPR}, 2020.

\bibitem{yi2016lift}
Kwang~Moo Yi, Eduard Trulls, Vincent Lepetit, and Pascal Fua.
\newblock Lift: Learned invariant feature transform.
\newblock In {\em ECCV}, 2016.

\bibitem{yi2018learning}
Kwang~Moo Yi, Eduard Trulls, Yuki Ono, Vincent Lepetit, Mathieu Salzmann, and
  Pascal Fua.
\newblock Learning to find good correspondences.
\newblock In {\em CVPR}, 2018.

\bibitem{yin2018geonet}
Zhichao Yin and Jianping Shi.
\newblock Geonet: Unsupervised learning of dense depth, optical flow and camera
  pose.
\newblock In {\em CVPR}, 2018.

\bibitem{yuan2020deepgmr}
Wentao Yuan, Benjamin Eckart, Kihwan Kim, Varun Jampani, Dieter Fox, and Jan
  Kautz.
\newblock Deepgmr: Learning latent gaussian mixture models for registration.
\newblock In {\em ECCV}, 2020.

\bibitem{zagoruyko2015learning}
Sergey Zagoruyko and Nikos Komodakis.
\newblock Learning to compare image patches via convolutional neural networks.
\newblock In {\em CVPR}, 2015.

\bibitem{zeng20163dmatch}
Andy Zeng, Shuran Song, Matthias Nie{\ss}ner, Matthew Fisher, Jianxiong Xiao,
  and Thomas Funkhouser.
\newblock 3dmatch: Learning local geometric descriptors from rgb-d
  reconstructions.
\newblock In {\em CVPR}, 2017.

\bibitem{zhang2020nerf++}
Kai Zhang, Gernot Riegler, Noah Snavely, and Vladlen Koltun.
\newblock Nerf++: Analyzing and improving neural radiance fields.
\newblock {\em arXiv}, 2020.

\bibitem{zhang1994iterative}
Zhengyou Zhang.
\newblock Iterative point matching for registration of free-form curves and
  surfaces.
\newblock {\em IJCV}, 1994.

\bibitem{zhang1995robust}
Zhengyou Zhang, Rachid Deriche, Olivier Faugeras, and Quang-Tuan Luong.
\newblock A robust technique for matching two uncalibrated images through the
  recovery of the unknown epipolar geometry.
\newblock {\em Artificial intelligence}, 1995.

\bibitem{open3d}
Qian-Yi Zhou, Jaesik Park, and Vladlen Koltun.
\newblock {Open3D}: {A} modern library for {3D} data processing.
\newblock {\em arXiv}, 2018.

\bibitem{zhou2017unsupervised}
Tinghui Zhou, Matthew Brown, Noah Snavely, and David~G Lowe.
\newblock Unsupervised learning of depth and ego-motion from video.
\newblock In {\em CVPR}, 2017.

\end{thebibliography}
}

\end{document}